\documentclass[letterpaper]{article}
\usepackage{aaai}
\usepackage{times}
\usepackage{helvet}
\usepackage{courier}
\usepackage{xcolor}
\usepackage{mathtools}
\usepackage{algorithm}
\usepackage{algpseudocode}
\usepackage{graphicx} 
\usepackage{url}
\usepackage{float}
\usepackage{listings}
\usepackage{mathdots}
\usepackage{soul}
\usepackage{caption}
\usepackage{subcaption}
\usepackage[export]{adjustbox}
\usepackage{natbib}

\lstdefinelanguage{PDDL}
{
  sensitive=false,    
  morecomment=[l]{;}, 
  alsoletter={:,-},   
  moredelim=**[is][\bf]{@}{@},
  morekeywords={
    define,domain,problem,not,and,or,when,forall,exists,either,
    :domain,:requirements,:types,:objects,:constants,
    :predicates,:action,:parameters,:precondition,:effect,
    :fluents,:primary-effect,:side-effect,:init,:goal,
    :strips,:adl,:equality,:typing,:conditional-effects,
    :negative-preconditions,:disjunctive-preconditions,
    :existential-preconditions,:universal-preconditions,:quantified-preconditions,
    :functions,assign,increase,decrease,scale-up,scale-down,
    :metric,minimize,maximize,
    :durative-actions,:duration-inequalities,:continuous-effects,
    :durative-action,:duration,:condition
  }
}

\usepackage{amsthm,amsmath,amsfonts}
\theoremstyle{definition}

\usepackage{cleveref}

\graphicspath{ {./artwork/src/} }

\begin{document}

\title{Bridging the Gap between Structural and Semantic Similarity in Diverse Planning}
\author{Mustafa F Abdelwahed\textsuperscript{1,2},
Joan Espasa\textsuperscript{1},
Alice Toniolo\textsuperscript{1},
Ian P. Gent\textsuperscript{1}\\
\textsuperscript{1}{University of St Andrews, School of Computer Science, UK}\\
\textsuperscript{2}{Helwan University, Faculty of Engineering, Egypt} \\
\{ma342, jea20, a.toniolo, ian.gent\}@st-andrews.ac.uk
}
\maketitle

\begin{abstract}
\begin{quote}
Diverse planning is the problem of finding multiple plans for a given problem specification, which is at the core of many real-world applications. 
For example, diverse planning is a critical piece for the efficiency of plan recognition systems when dealing with noisy and missing observations. Providing diverse solutions can also benefit situations where constraints are too expensive or impossible to model. 
Current diverse planners operate by generating multiple plans and then applying a selection procedure to extract diverse solutions using a similarity metric. 
Generally, current similarity metrics only consider the structural properties of the given plans. We argue that this approach is a limitation that sometimes prevents such metrics from capturing why two plans differ.
In this work, we propose two new domain-independent metrics which are able to capture relevant information on the difference between two given plans from a domain-dependent viewpoint. We showcase their utility in various situations where the currently used metrics fail to capture the similarity between plans, failing to capture some structural symmetries.
\end{quote}
\end{abstract}

\section{Introduction}
Given a set of elements, the maximum diversity problem~\citep{glover1977selecting} aims to find a subset of those elements that are maximally distanced apart in a metric space using a distance function.
In the context of combinatorial optimisation, sometimes it is not possible to model part of the problem because it is not available during the modelling phase. One could use the maximum diversity problem to obtain different solutions that help to overcome this drawback.
For example, multiple solutions for the cutting stock problem~\citep{haessler1991cutting} could provide a cutting procedure where the leftovers have a standard size, which can be potentially used later, thus providing a potentially better solution for a context that was not available during solving time.
Direct applications of the diversity problem are typically found in ensuring ethnic diversity~ \citep{bunzel1987diversity,kuo1993analyzing,mcconnell1988new} and generating diverse solutions for optimisation problems~\citep{hebrard2005finding,baste2022diversity}.
Applications that receive noisy or missing observations such as the malware detection~\citep{boddy2005course,sohrabi2013hypothesis}, applications that deal with simulation scenarios to provide risk management~\citep{sohrabi2018ai} or others that consider planning under pre-specified user preferences~\citep{myers1999generating,nguyen2012generating} either require or benefit from generating multiple solutions.

%
Current diverse automated planners generally solve the top-$k$ planning problem~\citep{riabov2014new}, a generalisation to optimal cost planning where the aim is to find the $k$ most optimal plans for a given planning task. There are two main strategies for solving the top-$k$ planning problem. The first uses a plan-forbid loop~\citep{katz2018novel,katz-sohrabi-aaai2020,katz2022needs}, using a planner to generate a solution and then reformulates the planning task, forcing the planner to avoid generating the same solution again. It keeps this procedure in a loop until the planner generates $k$ solutions. The second strategy uses symbolic search~\citep{speck2020symbolic}, which keeps exploring an abstraction of the search space until it finds a goal state. Then, it performs a backward search to generate a solution and continues generating the remaining $k-1$ solutions. 

After generating a fixed set of plans, top-$k$ planners perform a post-processing phase to select a diverse subset of the generated plans using a similarity metric. Currently, the available metrics are generally unable to differentiate between certain plan symmetries, such as plans with swapped resources. In addition to that, when a human needs to evaluate plans manually, the efficiency of the process can be hindered by either including or excluding symmetric plans in the selection phase.
By not considering those, both human effort and computational resources are wasted. Current metrics differentiate between two given plans using only the plan structural information~\citep{coman2011generating}. We instead propose to use semantic information extracted from the problem specification and the domain transition model.

The contributions of this paper are two novel domain-independent plan similarity metrics. The first metric measures similarity by comparing the plan reductions to partially ordered plans, while the second considers the order in which the two plans achieve the problem subgoals. Various case studies are then used to illustrate the new metrics' usefulness, showing that some structural symmetries can be captured and, therefore, the metrics better differentiate symmetrical plans.

\section{Diversity Metrics}
A planning task is a tuple of $\Pi=\langle F,A,I,G\rangle$, where $F$ is set of fluents, $A$ denotes a set of actions, $I$ represents the initial state and $G$ the goal formula.
A \emph{plan} is defined as a sequence of actions $a_1,a_2,\dots,a_n$ such that $a_i\in A$. Let $\mathcal{P}$ be a generated set of plans for a given planning problem, and $\pi$ one of its valid plans (i.e. $\pi\in\mathcal{P}$). 
%
Let $A(\pi)$ be the set of actions for plan $\pi$.
Let $\delta(\pi_a,\pi_b)\rightarrow{[0,1]}$ be a similarity function, which maps plans $\pi_a$ and $\pi_b$ to a bounded real number, where $0$ denotes the two plans are maximally different and $1$ indicates the two plans to be identical. Conversely, we define the dissimilarity between plans $\pi_a$ and $\pi_b$ as $D(\pi_a,\pi_b) = 1-\delta(\pi_a,\pi_b)$.

One of the first domain-independent approaches to generate diverse plans is proposed by~\citet{srivastava2007domain}, who suggested three distance functions used to select plans that are distanced apart from a given plan. 
Those similarity functions compared pairs of plans in terms of which actions are shared between them ($\delta_a$), the behaviours or states resulting after executing the actions in the plans ($\delta_s$), and the shared causal links, denoting which actions contribute to the goals being achieved ($\delta_c$). 
For a given plan, a causal link is a structure which links a producer action with a consumer action for a certain proposition. In other words, the producer's action has this proposition as an effect while the consumer has it as a precondition.
Causal links were initially used in the least commitment planning to find and validate partially-ordered plans~\citep{McAllesterR91,Weld94a}.
For any two given plans $\pi_a$ and $\pi_b$, these three similarity measures use the Jaccard measurement $\delta_{x}(\pi_a,\pi_b)=|A(\pi_a)\cap A(\pi_b)|/|A(\pi_b)\cup A(\pi_a)|$ where $x\in \{a,s,c\}$.
One more similarity metric is the uniqueness metric $\delta_{u}$~\cite{roberts2014evaluating}, defined as 
\begin{equation*}
\delta_u(\pi_a,\pi_b)={\begin{cases} 1,\ if\ \pi_a\setminus\pi_b=\emptyset  \\ 1,\ if\ \pi_a \subset \pi_b \\ 0,\ otherwise \end{cases}}
\label{equ:uniquness-measurment}
\end{equation*}
where plans are considered as sets of actions instead. The $\delta_u$ metric considers two plans unrelated if any action is presented in one plan but not included in the other. Its aim was to reduce the considered set of plans to those that are not subsumed after removing padded and permuted plans.

\begin{figure}
\begin{subfigure}[]{0.44\textwidth}
    \includegraphics[width=1.1\textwidth,left]{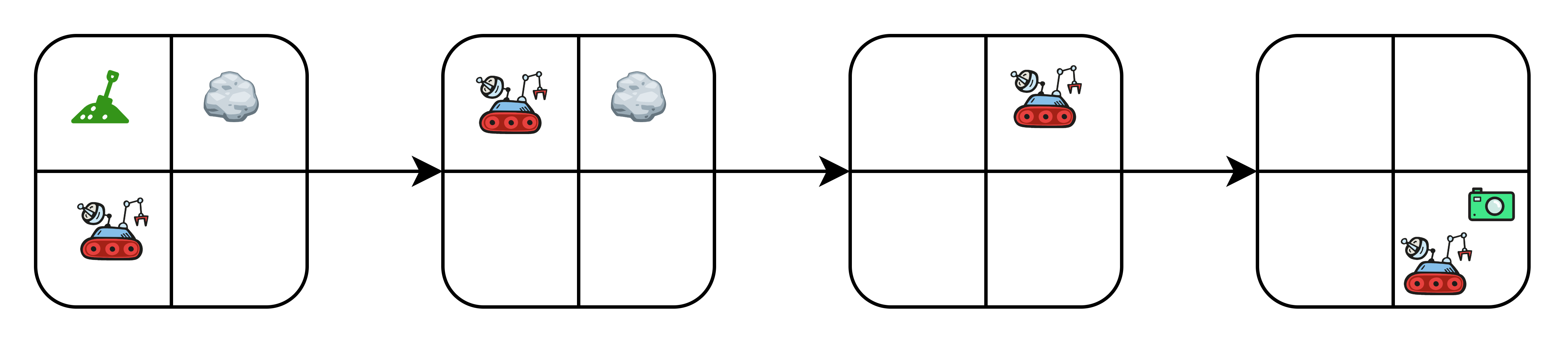}
    \caption{Plan I: the rover samples the soil followed by the rock and take an image.}
    \label{fig:rover-inst1-plan-1}
\end{subfigure}
\begin{subfigure}[]{0.44\textwidth}
    \includegraphics[width=1.1\textwidth,left]{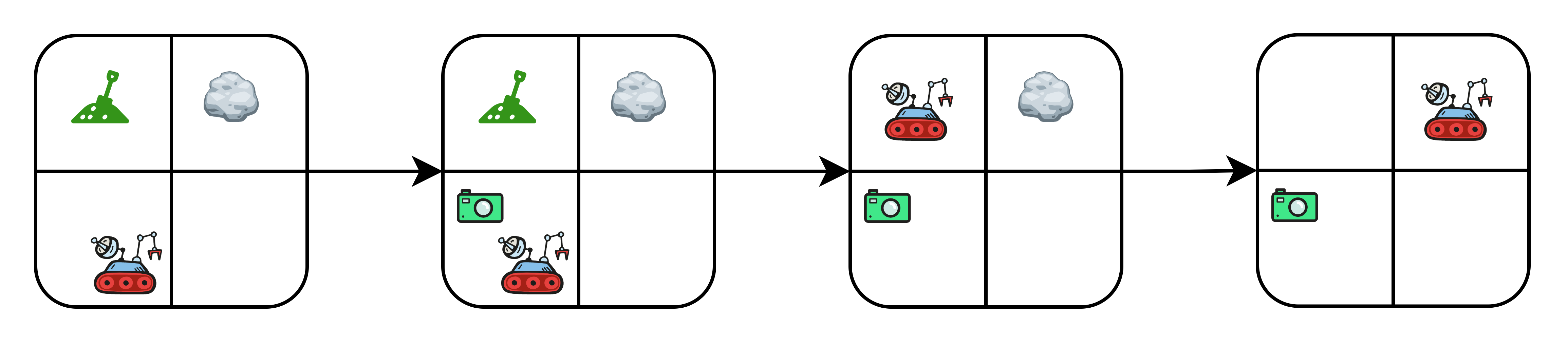}
    \caption{Plan II: the rover takes an image followed by the soil and then the rock.}
    \label{fig:rover-inst1-plan-2}
\end{subfigure}
\caption{Instance \#1 of the rover's planning problem}
\label{fig:rover-problem-initial-state}
\end{figure}

To further motivate the need for informative similarity metrics, consider the rover problem~\citep{IPC3} depicted in \Cref{fig:rover-problem-initial-state}. In this instance, the rover goals are:~$g_1$)~sample and communicate the soil, $g_2$)~sample and communicate the rock and finally, $g_3$)~send the image data.
When evaluating plans, a human modeller will need to understand why two plans differ and intuitively use various criteria to do so.
For example, a possible criterion would be to consider in what order the different goals are achieved. That is, the modeller could consider that a plan that achieves the subgoals in the order $g_1$, $g_2$, $g_3$ (\Cref{fig:rover-inst1-plan-1}) is inherently different from a plan that achieves them in the order $g_3$, $g_1$, $g_2$ (\Cref{fig:rover-inst1-plan-2}).
Another criterion could be to think about \emph{how} these subgoals are achieved. Similarly to HTN planning~\citep{erol1995hierarchical}, high-level actions or behaviours sometimes have to be divided in more than one basic action. 
Instead of focusing on the base actions, the redundant dependencies between parts of the plan are removed if total order plans are translated to partial order plans. The resulting partial order plans have then less noise to reason with.
Intuitively speaking, the metrics that should be used to discriminate between plans should use the same information as the human modeller.
%


%
Compared to~\citet{srivastava2007domain}, we argue that we can compare two plans concerning the flexibility of each plan (i.e. extracted partial-order plans) in addition to the sequence of subgoals achieved to reach a goal.

\subsection{Flexibility Metric} 
Our first proposed metric considers any two plans related if they share their extracted partial-ordered plans. Therefore the flexibility similarity metric is computed as the Jaccard measurement between the extracted partial order plans. The Jaccard measurement is selected as previous work~\citep{sohrabi2016finding} found that it tends to produce more diverse solutions. We define the flexibility metric as:
\begin{equation*}
\delta_{\text{flex}}(\pi_a,\pi_b)=\frac{|Pop(\pi_a)\cap Pop(\pi_b)|}{|Pop(\pi_a)\cup Pop(\pi_b)|}
\end{equation*}
where $Pop(\pi)$ is the partial order plan extracted from the total order plan $\pi$.
Much research covers how to extract partial-order plans~\citep{say2016mathematical,aghighi2017plan}. However, we used a simple approach~\citep{katz2018novel} to generate them, as it is computationally suitable for the post-processing phase. To clarify this, assume two plans $\pi_a$ and $\pi_b$, which are valid solutions for the rover problem mentioned in \Cref{fig:rover-problem-initial-state}. Each plan is a set of grounded actions, and each grounded action is mapped to a number and $$\pi_a=\{1,7,5,11,37,13,19,15,22,24\}$$ and $$\pi_{b}=\{1,7,5,11,13,19,15,22,24,37\}$$ For the action-based similarity metric, the value of 
\begin{equation*}
\delta_{a}(\pi_a,\pi_b)= \frac{|1,7,5,11,37,13,19,15,22,24|}{|1,7,5,11,13,19,15,22,24,37|} = 1
\end{equation*}
which suggests that the plans are identical. Our suggested metric aims to capture the dependency between those actions, 
\begin{multline*}
Pops(\pi_a) =\\ \{\{1,7\}, \{5\},\{11,37\}, \{13,19\}, \{15\}, \{22\}, \{24\}\}
\end{multline*}
and
\begin{multline*}
Pops(\pi_b)=\\ \{\{1,7\},  \{5,11\}, \{13,19\}, \{15\}, \{22\}, \{24,37\}\}
\end{multline*}
based on those extracted partial-order plans, we can compute the similarity as $\delta_{flex}= \frac{4}{9} = 0.4 $, thus indicating even though $\pi_a$ and $\pi_b$ have the same grounded actions still we can differentiate between them based on the dependency among actions in each plan.

\subsection{Subgoals Ordering Metric}

Our second proposed metric extracts the goal predicates from the problem's specification and considers each predicate as a subgoal, and then the metric aims to compare two plans by comparing in what order and when the sub-tasks in the task are achieved. Since subgoals are effectively landmarks, this metric relates to the landmark-based distance function presented by~\citet{bryce2014landmark}. However, our metric is computationally cheap compared to the expensive computation for landmarks, as we can infer it directly from the problem specification. 
This second metric normalises the hamming distance 
between the subgoals sequences and the max subgoal sequence as follows:
\begin{equation*}
\delta_{\text{sgo}}(\pi_a,\pi_b)=1-\frac{HDist(SubGoals(\pi_a), SubGoals(\pi_b))}{max(SubGoals(\pi_a),SubGoals(\pi_b))}
\end{equation*}


where $SubGoals$, defined in \Cref{alg:subgoals}, is a function that receives a total order plan and returns a string encoding the order in which subgoals are achieved. 
$HDist(s_{\pi_a}, s_{\pi_b})\rightarrow\mathbb{R}$ be a function that computes the hamming distance between the subgoal sequences resulting from the $SubGoals$ function applied to a given pair of plans. 
Note that the return value of the $SubGoals$ function preserves the step in which the subgoals are achieved, and therefore the $HDist$ function is considering this information. Since $\delta_{sgo}$ is expected to return a normalised value between $[0,1]$, we normalise the Hamming distance by the maximum string length using the $max(a, b)\rightarrow\mathbb{N}$ function.
\begin{algorithm}[H]
\caption{$SubGoals$}\label{alg:subgoals}
\begin{algorithmic}[1]
\Require $\pi$: Plan, PI: Problem Instance
\Ensure $seq$
\State $subgoalLetter \gets GetEncodedSubgoals(PI)$\label{alg-line:get-letters}
\State $seq \gets ``"$\label{alg-line:empty-string}
\State $state \gets GetInitialState(PI)$\label{alg-line:initial-state}
\For{$a \in \pi$}
\State $state \gets PerformAction(state, a)$ \label{alg-line:start-loop}
\State $sg \gets GetSubGoal(state, PI)$
\If{$sg \neq X$}
    \State $seq \gets AppendTo(seq, subgoalLetter[sg])$
\Else
    \State $seq \gets AppendTo(seq, ``X")$
\EndIf
\EndFor \label{alg-line:end-loop}
\end{algorithmic}
\end{algorithm}

\Cref{alg:subgoals} starts with getting an encoded character map of the available subgoals in the provided problem instance 
(\Cref{alg-line:get-letters}) and then creates an empty string. It then gets the initial state to simulate the actions in $\pi$ 
(\Cref{alg-line:empty-string}-\Cref{alg-line:initial-state}). Afterwards, $SubGoals$ simulates each action and checks the successor state to see if it contains an achieved subgoal. 
If it achieves one, it appends its encoded letter; otherwise, it appends $X$, which indicates no subgoals achieved (\Cref{alg-line:start-loop}-\Cref{alg-line:end-loop}). 

To illustrate both the intuition behind this metric and how the $SubGoals$ function works, let us consider the rover problem presented in \Cref{fig:rover-problem-initial-state}. The rover aimed to sample and communicate the soil, rock and image data. The $SubGoals$ function encodes those subgoals into characters: the communicating soil subgoal is encoded into the letter $A$, the rock subgoal is encoded with the letter $B$ and the image data with the letter $C$.

Assume plans $\pi_a$ and $\pi_b$ solve the problem in different subgoals sequences. More concretely, $\pi_a$ sends the rock data after three actions, followed by five actions to send the soil data and finally sends the image data after two actions. On the other hand, $\pi_b$ sends the image data after four actions, followed by one action to send the rock data, and finally sends the soil data after five actions. 
Based on our encoding characters map, $SubGoals(\pi_a)$ would consider $BAC$ while $SubGoals(\pi_b)$ generates $CBA$. Moreover, we want to include information about when those subgoals got achieved. Therefore, to account for the timestep $SubGoals$ uses the letter $X$ to represent states with no subgoals accomplished, returning $SubGoals(\pi_a)=``XXBXXXXAXC"$ and $SubGoals(\pi_b)=``XXXCBXXXXA"$. Based on these encoded plans, we can now compute the hamming distance, which will be $HDist(SubGoals(\pi_a), SubGoals(\pi_b))=5$. Note that this value encodes the difference in subgoal ordering and their positions. The final $\delta_{sgo}$ value would then be $\frac{5}{10}$ where \emph{10} is the number of states (i.e. the maximum encoded string length).
%

\section{Case Studies}

This section illustrates the behaviour of the proposed similarity metrics. To showcase our proposed metrics, we have extended the \emph{Diverse-score} software\footnote{\url{https://github.com/IBM/diversescore}} to include our suggested metrics\footnote{\url{https://github.com/MFaisalZaki/PAIR2023-Semantic-Similarity-Metrics}}. We used SYM-K~\citep{speck2020symbolic} to generate all optimal plans for a series of well-known domains. Afterwards, we selected two plans that show the strengths and weakness of the suggested metrics when compared to the following currently available metrics: stability ($\delta_a$), states ($\delta_s$), and uniqueness ($\delta_u$). 
We will not use $\delta_c$ in our comparisons, as previous research~\citep{srivastava2007domain} concluded that $\delta_{a}$ produced more diverse plans compared to $\delta_{s}$ and $\delta_{c}$. However, we will use $\delta_{a}$ and $\delta_{s}$ in our comparisons, as $\delta_{a}$ was considered the best metric in~\citep{srivastava2007domain} and as $\delta_{flex}$, uses the information of what actions appear in the plan. $\delta_s$, similarly to $\delta_{sgo}$, considers the state trajectories of the plan. To examine the computational time consumed by our similarity metrics, we have added in each table the execution time in seconds when computing the scores for those plans.

\subsection {Depots Planning Problem}
The Depot domain was introduced in the third IPC and combined two well-known problems: BlocksWorld and Logistics.  
The possible actions are to stack and unstack crates using hoists and to move these crates between different locations using trucks. It is characterised by having significant goal interaction.

\begin{figure}[t]
\begin{small}
\begin{lstlisting}[language=PDDL,frame=single,numbers=left,stepnumber=1,firstnumber=1]
(lift hoist0 crate0 pallet0 depot0)
(load hoist0 crate0 @truck0@ depot0)
(lift hoist2 crate2 crate1 distributor1)
(drive @truck0@ depot0 distributor1)
(load hoist2 crate2 @truck0@ distributor1)
(lift hoist2 crate1 pallet2 distributor1)
(load hoist2 crate1 @truck0@ distributor1)
(unload hoist2 crate0 @truck0@ distributor1)
(drive @truck0@ distributor1 depot0)
(drop hoist2 crate0 pallet2 distributor1)
(unload hoist0 crate2 @truck0@ depot0)
(drive @truck0@ depot0 distributor0)
(drop hoist0 crate2 pallet0 depot0)
(unload hoist1 crate1 @truck0@ distributor0)
(drop hoist1 crate1 crate3 distributor0)
; cost = 15 (unit cost)
\end{lstlisting}
\begin{lstlisting}[language=PDDL,frame=single,numbers=left,stepnumber=1,firstnumber=1,firstnumber=1]
(lift hoist0 crate0 pallet0 depot0)
(load hoist0 crate0 @truck1@ depot0)
(lift hoist2 crate2 crate1 distributor1)
(drive @truck1@ depot0 distributor1)
(load hoist2 crate2 @truck1@ distributor1)
(lift hoist2 crate1 pallet2 distributor1)
(load hoist2 crate1 @truck1@ distributor1)
(unload hoist2 crate0 @truck1@ distributor1)
(drive @truck1@ distributor1 depot0)
(drop hoist2 crate0 pallet2 distributor1)
(unload hoist0 crate2 @truck1@ depot0)
(drive @truck1@ depot0 distributor0)
(drop hoist0 crate2 pallet0 depot0)
(unload hoist1 crate1 @truck1@ distributor0)
(drop hoist1 crate1 crate3 distributor0)
; cost = 15 (unit cost)
\end{lstlisting}
\end{small}
\caption{Considered symmetrical plans for the depots planning problem.}
\label{plan:depots-symmterical-1}
\end{figure} 


\begin{table}
\centering
\begin{tabular}{|c|c|c|c|}
\hline
\textbf{\#} & \textbf{Metric} & \textbf{Similarity Score} & \textbf{Computation time} \\ \hline
1           & $\delta_{sgo}$  & 1                         & 0.953                              \\ \hline
2           & $\delta_{s}$    & 0.79                      & 0.913                              \\ \hline
3           & $\delta_{a}$    & 0.25                      & 0.963                              \\ \hline
4           & $\delta_{flex}$ & 0.14                      & 0.988                              \\ \hline
5           & $\delta_{u}$    & 0                         & 0.903                              \\ \hline
\end{tabular}
\caption{Similarity score of the symmetrical depot plans and the total computation time in milliseconds.}
\label{tbl:similarity-symmterical-score-depots}
\end{table}

We used instance\#2 from the IPC-2002 as a case study. \Cref{plan:depots-symmterical-1} shows two plans to solve the instance that we consider symmetrical, as one plan used \texttt{truck0} to solve the planning task while the second used \texttt{truck1}. \Cref{tbl:similarity-symmterical-score-depots} shows the values of the computed metrics between these two plans. Unlike $\delta_{a}$, $\delta_{sgo}$ was able to capture the plan symmetry. Note that $\delta_{s}$ could  unintendedly capture the similarity to some extent if the number of state variables related to the trucks were a small enough set when compared to the whole set of state variables. Since $\delta_{flex}$ and $\delta_{u}$ use grounded actions to compare similarity, they clearly did not capture symmetrical information. 

\Cref{plan:depots-meaningful-1} shows two plans which are almost identical in structure except for the order in which the subgoals are accomplished. We will consider those plans to be dissimilar, as even if both plans accomplish the goal if one were to visualise the execution of both plans in parallel, one would intuitively consider them sufficiently different.
%
%
More concretely, the subgoals are achieved by the \emph{drop} actions in \Cref{plan:depots-meaningful-1}, marked in bold. 

%
%
\begin{figure}[t]
\begin{small}
\begin{lstlisting}[language=PDDL,frame=single,numbers=left,stepnumber=1,firstnumber=1]
(lift hoist0 crate0 pallet0 depot0)
(load hoist0 crate0 truck0 depot0)
(lift hoist2 crate2 crate1 distributor1)
(drive truck0 depot0 distributor1)
(load hoist2 crate2 truck0 distributor1)
(lift hoist2 crate1 pallet2 distributor1)
(load hoist2 crate1 truck0 distributor1)
(unload hoist2 crate0 truck0 distributor1)
(drive truck0 distributor1 depot0)
@(drop hoist2 crate0 pallet2 distributor1)@
(unload hoist0 crate2 truck0 depot0)
(drive truck0 depot0 distributor0)
@(drop hoist0 crate2 pallet0 depot0)@
(unload hoist1 crate1 truck0 distributor0)
@(drop hoist1 crate1 crate3 distributor0)@
; cost = 15 (unit cost)
\end{lstlisting}
\begin{lstlisting}[language=PDDL,frame=single,numbers=left,stepnumber=1,firstnumber=1]
(lift hoist0 crate0 pallet0 depot0)
(load hoist0 crate0 truck0 depot0)
(lift hoist2 crate2 crate1 distributor1)
(drive truck0 depot0 distributor1)
(load hoist2 crate2 truck0 distributor1)
(lift hoist2 crate1 pallet2 distributor1)
(load hoist2 crate1 truck0 distributor1)
(unload hoist2 crate0 truck0 distributor1)
(drive truck0 distributor1 depot0)
(unload hoist0 crate2 truck0 depot0)
(drive truck0 depot0 distributor0)
@(drop hoist0 crate2 pallet0 depot0)@
(unload hoist1 crate1 truck0 distributor0)
@(drop hoist1 crate1 crate3 distributor0)@
@(drop hoist2 crate0 pallet2 distributor1)@
; cost = 15 (unit cost)
\end{lstlisting}
\end{small}
\caption{Semantically different plans for the depots planning problem.}
\label{plan:depots-meaningful-1}
\end{figure} 


\begin{table}
\centering
\begin{tabular}{|c|c|c|c|}
\hline
\textbf{\#} & \textbf{Metric} & \textbf{Similarity Score} & \textbf{Computation time} \\ \hline
1           & $\delta_{u}$    & 1                         & 0.960                              \\ \hline
2           & $\delta_{a}$    & 1                         & 0.920                              \\ \hline
3           & $\delta_{s}$    & 0.81                      & 0.934                              \\ \hline
4           & $\delta_{flex}$ & 0.714                     & 0.962                              \\ \hline
5           & $\delta_{sgo}$  & 0.67                      & 0.928                              \\ \hline

\end{tabular}
\caption{Similarity score for the semantically different depot plans and the total computation time in milliseconds.}
\label{tbl:similarity-meaningful-score-depots}
\end{table}

\Cref{tbl:similarity-meaningful-score-depots} lists the similarity score of each metric for those two plans. Notice that $\delta_{sgo}$ was able to differentiate between those two plans, unlike $\delta_{a}$, which considers them identical plans. This is due to $\delta_{sgo}$ considering the goal specification, which is not the case in all other similarity metrics. 

\subsection{Satellite Planning Problem}

\begin{figure}
    \begin{subfigure}[]{0.45\textwidth}
        \centering
        \includegraphics[width=1.04\textwidth,left]{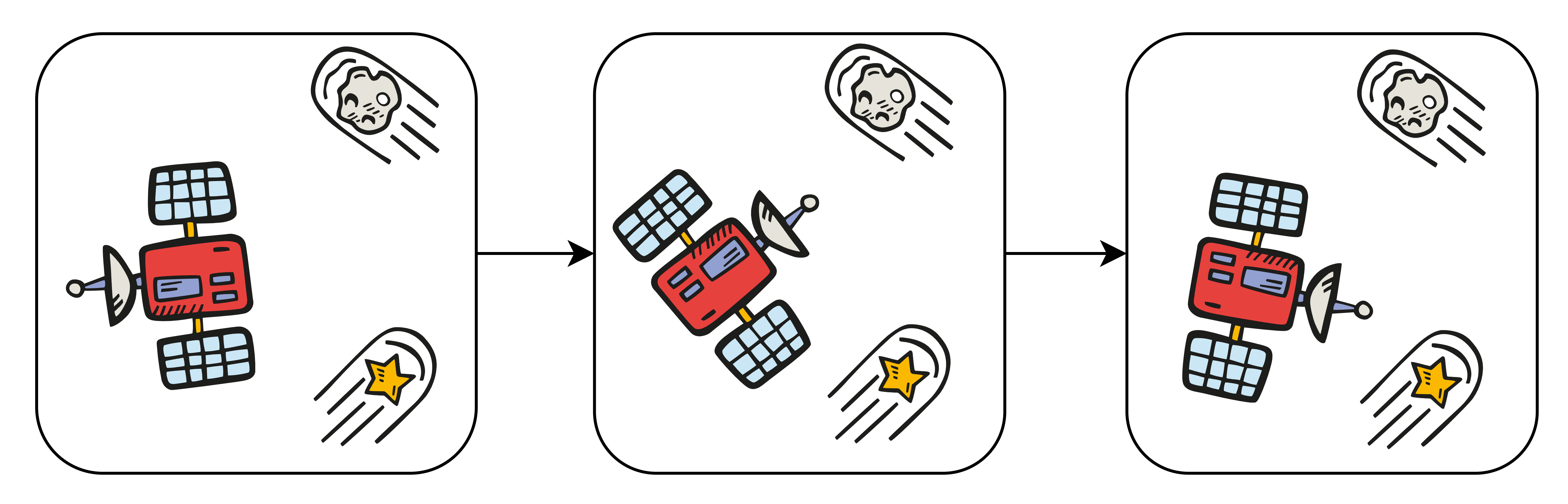}
        \caption{Plan I: satellite takes an image for \texttt{phenomenon6} followed by \texttt{star5}. }
    \end{subfigure}

    \begin{subfigure}[]{0.45\textwidth}
        \centering
        \includegraphics[width=1.04\textwidth,left]{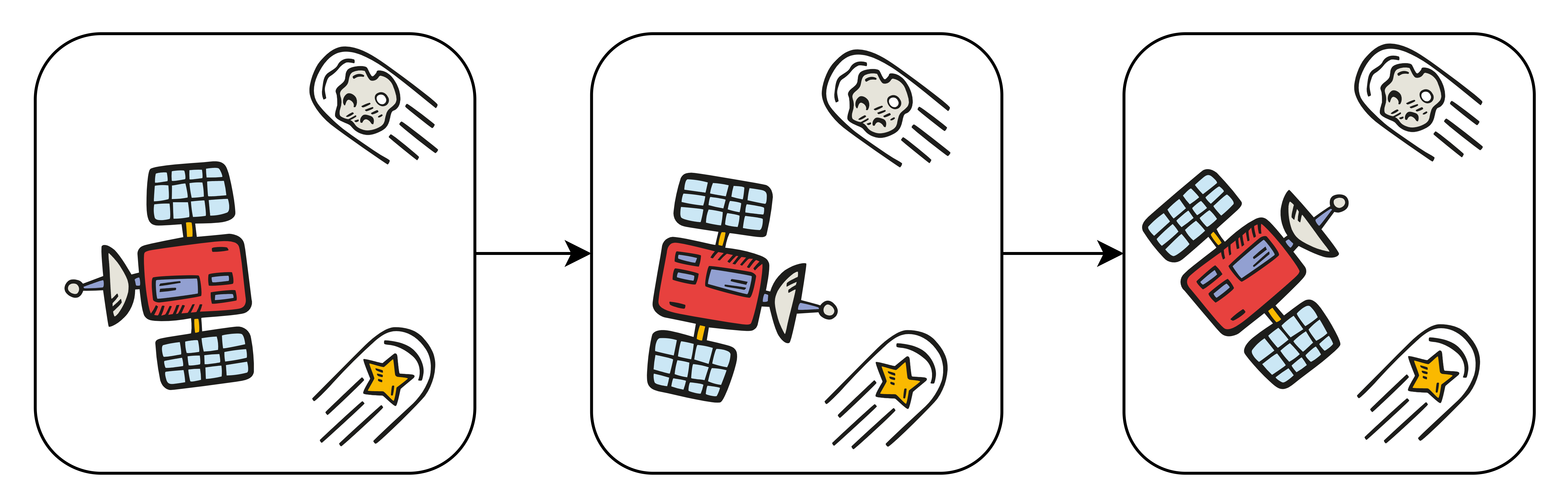}
        \caption{Plan II: satellite takes an image for \texttt{star5} followed by \texttt{phenomenon6}. }
    \end{subfigure}
    \caption{Reversed order plans for the satellite planning problem.}
    \label{plan:satellite-reversed-1}
\end{figure}

The satellite domain~\cite{IPC3} involves scheduling and coordinating multiple satellites to perform various observation tasks. Each satellite may have different capabilities, such as sensors with different resolutions or ranges. Besides, there may also be implicit constraints on which tasks can be performed by each satellite. 
%
%
If we do not consider unnecessary actions or cycles in the plan, the only possible variation in this problem is the order in which the different actions need to be executed. \Cref{plan:satellite-reversed-1} shows two different plans, where one plan starts with taking an image for \texttt{phenomenon6} followed by \texttt{star5}. The second plan takes an image for \texttt{star5} followed by \texttt{phenomenon6} in this order. Note in \Cref{plan:satellite-reversed-1} that those two plans share the same structure and order except for the swapped phenomena/star. 
%
%


\begin{table}
\centering
\begin{tabular}{|c|c|c|c|}
\hline
\textbf{\#} & \textbf{Metric} & \textbf{Diverse Score} & \textbf{Computation time } \\ \hline
1           & $\delta_{sgo}$  & 0.78                   & 0.44                               \\ \hline
2           & $\delta_{s}$    & 0.67                   & 0.45                               \\ \hline
3           & $\delta_{a}$    & 0.55                   & 0.42                               \\ \hline
4           & $\delta_{flex}$ & 0.46                   & 0.44                               \\ \hline
5           & $\delta_{u}$    & 0                      & 0.43                               \\ \hline
\end{tabular}
\caption{Diverse score of the selected satellite plans and the total computation time in milliseconds.}
\label{tbl-satellite-problem}
\end{table}
%
Similar to \Cref{tbl:similarity-meaningful-score-depots}, \Cref{tbl-satellite-problem} presents the similarity scores for the satellite plans. Plans presented in \Cref{plan:satellite-reversed-1} are similar in some sense according to human intuition, and $\delta_{a}$, $\delta_{s}$ and $\delta_{sgo}$ did an excellent job in detecting the similarity between them. 
The similarity between $\delta_{sgo}$ and $\delta_{s}$ was higher compared to $\delta_{a}$. Due to the $\delta_{sgo}$ focus on subgoals only and disregards the structure, whereas $\delta_{s}$ utilises state variables to compare plans that consist of subgoals treated as fluents and not given higher influence on the plan. Unfortunately, $\delta_{flex}$ and $\delta_u$ failed to compete with the other metrics; it seems that capturing the dependency between actions or searching for unique actions only will not be sufficient to differentiate between plans.

%
%
\subsection {Zenotravel Planning Problem}
\begin{figure*}
    \centering
    \begin{subfigure}[]{0.8\textwidth}
        \centering
        \includegraphics[scale=0.04]{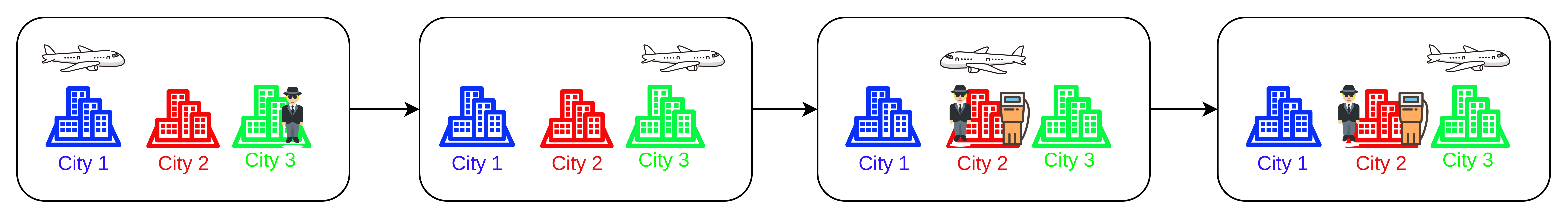}
        \caption{Plan I: picks the passenger, then refuel after debarking the passenger at his destination.}
    \end{subfigure}

    \begin{subfigure}[]{0.8\textwidth}
        \centering
        \includegraphics[scale=0.04]{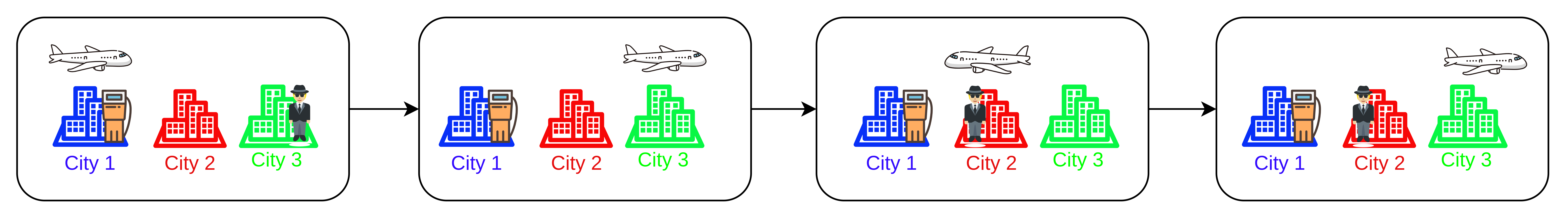}
        \caption{Plan II: first refuel the plane and then fly to transport the passenger to his destination.}
    \end{subfigure}
    
    \caption{Selected plans for the zenotravel planning problem.}
    \label{plan:zenotravel-reversed-1}
\end{figure*}

In this transportation domain, people are moved around in aircraft. The key part of the domain is that when zooming, more fuel is consumed than when using a regular speed when flying. \Cref{plan:zenotravel-reversed-1} shows two plans, where the \emph{refuel} action is executed either at the beginning (Line 1 in the former) or at the end of the plan (Line 5 in the latter).


\begin{table}
\centering
\begin{tabular}{|c|c|c|c|}
\hline 
\textbf{\#} & \textbf{Metric} & \textbf{Diverse Score} & \textbf{Computation time } \\ \hline
1           & $\delta_{sgo}$  & 0.67                   & 0.58                               \\ \hline
2           & $\delta_{s}$    & 0.56                   & 0.58                               \\ \hline
3           & $\delta_{a}$    & 0.33                   & 0.58                               \\ \hline
4           & $\delta_{flex}$ & 0.22                   & 0.58                               \\ \hline
5           & $\delta_{u}$    & 0                      & 0.58                               \\ \hline
\end{tabular}
\caption{Diverse score of the selected zenotravel plans and the total computation time in milliseconds.}
\label{tbl-zenotravel-problem}
\end{table}

%
%
\Cref{tbl-zenotravel-problem} shows the similarity scores computed for those plans. This confirms the same findings in the satellite case study except on a smaller plan length with different grounded actions and similar execution on a human-level intuition. The only noticeable difference between those two plans is when a plan \texttt{refuel} the plane, such a difference had no noticeable effect except for the sequence of visiting the cities. Suchlike situations show the need to improve the $\delta_{sgo}$ by accounting for landmarks.

%
%
%
\section{Discussion and Future Work}
The primary contribution behind this paper is extending \citep{srivastava2007domain}'s perspective to compare two plans together by presenting two similarity metrics that better approximate a domain's modeller intuition.
The currently used metrics consider the similarity between any two plans by measuring the number of shared grounded actions between them. In the depot's case study, we highlighted a subjective question of whether symmetrical plans should be considered identical. We argue that two plans doing the same thing in the same order but using different resources are identical, while on a plan structural level, those plans are different. Since some domains could contain various symmetrical plans, the submitted metrics must account for this and try to distinguish between two plans on a higher logic level. Therefore, $\delta_{flex}$ tried to reason about the common dependencies between actions by converting a total order plan into a partially ordered one and then comparing those sets of partial plans. Unfortunately, $\delta_{flex}$ did not perform as expected, and such a degraded performance could have been influenced by multiple factors, such as the algorithms used for extracting partial order plans, the plan length and the domain nature that could allow parallel plan execution to some extent. On the contrary, $\delta_{sgo}$ showed a high potential when comparing symmetrical plans since $\delta_{sgo}$ compares how and when two given plans achieve their goals by checking the order of the subgoals predicate extracted from the problem specification. 

The other case studies, satellite and zenotravel, showed the details $\delta_{sgo}$ could capture between any two plans. Thus, we believe  it could arguably match the domain modeller's intuition when comparing plans. Such an ability is highlighted in zenotravel, where the provided plans showed an almost identical sequence of execution but differed in one action. For the current similarity metrics $\delta_{a}$, $\delta_{flex}$ and $\delta_{u}$ are inferring their information from the plan's structure which was unsuccessful in capturing the similarity between those plans, unlike the $\delta_{sgo}$ and $\delta_{s}$ which inferred their conclusions from state variables. Considering and comparing state variables can capture relevant information when checking for similarity, and in addition considering subgoals sequences makes the metric more resilient to symmetry. 

The presented metrics can now be integrated into diverse planning and plan recognition-related applications. Still, many further improvements can be made to the presented similarity metrics. One possible improvement for $\delta_{sgo}$ is considering landmarks in addition to the subgoal predicates. 
Another improvement related to the domain modeller's intuition is considering aggregating multiple similarity metrics together, such as assembling $\delta_{sgo}$ with $\delta_{a}$, which holds the information of having a plan that solves the planning problem in a different sequence with different grounded actions.

\bibliographystyle{aaai}
\bibliography{ref}

\begin{thebibliography}{}

\bibitem[\protect\citeauthoryear{Aghighi and
  B{\"a}ckstr{\"o}m}{2017}]{aghighi2017plan}
Aghighi, M., and B{\"a}ckstr{\"o}m, C.
\newblock 2017.
\newblock Plan reordering and parallel execution—a parameterized complexity
  view.
\newblock In {\em Thirty-First AAAI Conference on Artificial Intelligence}.

\bibitem[\protect\citeauthoryear{Baste \bgroup et al\mbox.\egroup
  }{2022}]{baste2022diversity}
Baste, J.; Fellows, M.~R.; Jaffke, L.; Masa{\v{r}}{\'\i}k, T.;
  de~Oliveira~Oliveira, M.; Philip, G.; and Rosamond, F.~A.
\newblock 2022.
\newblock Diversity of solutions: An exploration through the lens of
  fixed-parameter tractability theory.
\newblock {\em Artificial Intelligence} 303:103644.

\bibitem[\protect\citeauthoryear{Boddy \bgroup et al\mbox.\egroup
  }{2005}]{boddy2005course}
Boddy, M.~S.; Gohde, J.; Haigh, T.; and Harp, S.~A.
\newblock 2005.
\newblock Course of action generation for cyber security using classical
  planning.
\newblock In {\em ICAPS},  12--21.

\bibitem[\protect\citeauthoryear{Bryce}{2014}]{bryce2014landmark}
Bryce, D.
\newblock 2014.
\newblock Landmark-based plan distance measures for diverse planning.
\newblock In {\em Proceedings of the International Conference on Automated
  Planning and Scheduling}, volume~24,  56--64.

\bibitem[\protect\citeauthoryear{Bunzel and Au}{1987}]{bunzel1987diversity}
Bunzel, J.~H., and Au, J.~K.
\newblock 1987.
\newblock Diversity or discrimination?-asian americans in college.
\newblock {\em The Public Interest} 87:49.

\bibitem[\protect\citeauthoryear{Coman and
  Munoz-Avila}{2011}]{coman2011generating}
Coman, A., and Munoz-Avila, H.
\newblock 2011.
\newblock Generating diverse plans using quantitative and qualitative plan
  distance metrics.
\newblock In {\em Proceedings of the AAAI Conference on Artificial
  Intelligence}, volume~25,  946--951.

\bibitem[\protect\citeauthoryear{Erol}{1995}]{erol1995hierarchical}
Erol, K.
\newblock 1995.
\newblock {\em Hierarchical task network planning: formalization, analysis, and
  implementation}.
\newblock University of Maryland, College Park.

\bibitem[\protect\citeauthoryear{Glover, Hersh, and
  McMillan}{1977}]{glover1977selecting}
Glover, F.; Hersh, G.; and McMillan, C.
\newblock 1977.
\newblock Selecting subsets of maximum diversity, ms.
\newblock Technical report, IS Report.

\bibitem[\protect\citeauthoryear{Haessler and
  Sweeney}{1991}]{haessler1991cutting}
Haessler, R.~W., and Sweeney, P.~E.
\newblock 1991.
\newblock Cutting stock problems and solution procedures.
\newblock {\em European Journal of Operational Research} 54(2):141--150.

\bibitem[\protect\citeauthoryear{Hebrard \bgroup et al\mbox.\egroup
  }{2005}]{hebrard2005finding}
Hebrard, E.; Hnich, B.; O'Sullivan, B.; and Walsh, T.
\newblock 2005.
\newblock Finding diverse and similar solutions in constraint programming.
\newblock In {\em AAAI}, volume~5,  372--377.

\bibitem[\protect\citeauthoryear{Katz and
  Sohrabi}{2020}]{katz-sohrabi-aaai2020}
Katz, M., and Sohrabi, S.
\newblock 2020.
\newblock Reshaping diverse planning.
\newblock In {\em Proceedings of the Thirty-Fourth {AAAI} Conference on
  Artificial Intelligence ({AAAI} 2020)},  9892--9899.
\newblock {AAAI} Press.

\bibitem[\protect\citeauthoryear{Katz and Sohrabi}{2022}]{katz2022needs}
Katz, M., and Sohrabi, S.
\newblock 2022.
\newblock Who needs these operators anyway: Top quality planning with operator
  subset criteria.
\newblock In {\em Proceedings of the International Conference on Automated
  Planning and Scheduling}, volume~32,  179--183.

\bibitem[\protect\citeauthoryear{Katz \bgroup et al\mbox.\egroup
  }{2018}]{katz2018novel}
Katz, M.; Sohrabi, S.; Udrea, O.; and Winterer, D.
\newblock 2018.
\newblock A novel iterative approach to top-k planning.
\newblock In {\em Twenty-Eighth International Conference on Automated Planning
  and Scheduling}.

\bibitem[\protect\citeauthoryear{Kuo, Glover, and
  Dhir}{1993}]{kuo1993analyzing}
Kuo, C.-C.; Glover, F.; and Dhir, K.~S.
\newblock 1993.
\newblock Analyzing and modeling the maximum diversity problem by zero-one
  programming.
\newblock {\em Decision Sciences} 24(6):1171--1185.

\bibitem[\protect\citeauthoryear{Long and Fox}{2003}]{IPC3}
Long, D., and Fox, M.
\newblock 2003.
\newblock The 3rd international planning competition: Results and analysis.
\newblock {\em J. Artif. Intell. Res.} 20:1--59.

\bibitem[\protect\citeauthoryear{McAllester and
  Rosenblitt}{1991}]{McAllesterR91}
McAllester, D.~A., and Rosenblitt, D.
\newblock 1991.
\newblock {Systematic Nonlinear Planning}.
\newblock In Dean, T.~L., and McKeown, K.~R., eds., {\em Proceedings of the 9th
  National Conference on Artificial Intelligence, Anaheim, CA, USA, July 14-19,
  1991, Volume 2},  634--639.
\newblock {AAAI} Press / The {MIT} Press.

\bibitem[\protect\citeauthoryear{McConnell}{1988}]{mcconnell1988new}
McConnell, S.
\newblock 1988.
\newblock The new battle over immigration.
\newblock {\em Fortune} 117(10):89.

\bibitem[\protect\citeauthoryear{Myers and Lee}{1999}]{myers1999generating}
Myers, K.~L., and Lee, T.~J.
\newblock 1999.
\newblock Generating qualitatively different plans through metatheoretic
  biases.
\newblock In {\em AAAI/IAAI},  570--576.

\bibitem[\protect\citeauthoryear{Nguyen \bgroup et al\mbox.\egroup
  }{2012}]{nguyen2012generating}
Nguyen, T.~A.; Do, M.; Gerevini, A.~E.; Serina, I.; Srivastava, B.; and
  Kambhampati, S.
\newblock 2012.
\newblock Generating diverse plans to handle unknown and partially known user
  preferences.
\newblock {\em Artificial Intelligence} 190:1--31.

\bibitem[\protect\citeauthoryear{Riabov, Sohrabi, and
  Udrea}{2014}]{riabov2014new}
Riabov, A.; Sohrabi, S.; and Udrea, O.
\newblock 2014.
\newblock New algorithms for the top-k planning problem.
\newblock In {\em Proceedings of the scheduling and planning applications
  workshop (spark) at the 24th international conference on automated planning
  and scheduling (icaps)},  10--16.

\bibitem[\protect\citeauthoryear{Roberts, Howe, and
  Ray}{2014}]{roberts2014evaluating}
Roberts, M.; Howe, A.; and Ray, I.
\newblock 2014.
\newblock Evaluating diversity in classical planning.
\newblock In {\em Proceedings of the International Conference on Automated
  Planning and Scheduling}, volume~24,  253--261.

\bibitem[\protect\citeauthoryear{Say, Cire, and
  Beck}{2016}]{say2016mathematical}
Say, B.; Cire, A.~A.; and Beck, J.~C.
\newblock 2016.
\newblock Mathematical programming models for optimizing partial-order plan
  flexibility.
\newblock In {\em Proceedings of the Twenty-second European Conference on
  Artificial Intelligence},  1044--1052.

\bibitem[\protect\citeauthoryear{Sohrabi \bgroup et al\mbox.\egroup
  }{2016}]{sohrabi2016finding}
Sohrabi, S.; Riabov, A.~V.; Udrea, O.; and Hassanzadeh, O.
\newblock 2016.
\newblock Finding diverse high-quality plans for hypothesis generation.
\newblock In {\em ECAI 2016}. IOS Press.
\newblock  1581--1582.

\bibitem[\protect\citeauthoryear{Sohrabi \bgroup et al\mbox.\egroup
  }{2018}]{sohrabi2018ai}
Sohrabi, S.; Riabov, A.; Katz, M.; and Udrea, O.
\newblock 2018.
\newblock An ai planning solution to scenario generation for enterprise risk
  management.
\newblock In {\em Proceedings of the AAAI Conference on Artificial
  Intelligence}, volume~32.

\bibitem[\protect\citeauthoryear{Sohrabi, Udrea, and
  Riabov}{2013}]{sohrabi2013hypothesis}
Sohrabi, S.; Udrea, O.; and Riabov, A.
\newblock 2013.
\newblock Hypothesis exploration for malware detection using planning.
\newblock In {\em Proceedings of the AAAI Conference on Artificial
  Intelligence}, volume~27,  883--889.

\bibitem[\protect\citeauthoryear{Speck, Mattm{\"u}ller, and
  Nebel}{2020}]{speck2020symbolic}
Speck, D.; Mattm{\"u}ller, R.; and Nebel, B.
\newblock 2020.
\newblock Symbolic top-k planning.
\newblock In {\em Proceedings of the AAAI Conference on Artificial
  Intelligence}, volume~34,  9967--9974.

\bibitem[\protect\citeauthoryear{Srivastava \bgroup et al\mbox.\egroup
  }{2007}]{srivastava2007domain}
Srivastava, B.; Nguyen, T.~A.; Gerevini, A.; Kambhampati, S.; Do, M.~B.; and
  Serina, I.
\newblock 2007.
\newblock Domain independent approaches for finding diverse plans.
\newblock In {\em IJCAI},  2016--2022.

\bibitem[\protect\citeauthoryear{Weld}{1994}]{Weld94a}
Weld, D.~S.
\newblock 1994.
\newblock An introduction to least commitment planning.
\newblock {\em {AI} Mag.} 15(4):27--61.

\end{thebibliography}

\end{document}